\documentclass[conference]{IEEEtran}
\IEEEoverridecommandlockouts
\usepackage{cite}
\usepackage{amsmath,amssymb,amsfonts}
\usepackage{algorithmic}
\usepackage{graphicx}
\usepackage{textcomp}
\usepackage{xcolor}
\def\BibTeX{{\rm B\kern-.05em{\sc i\kern-.025em b}\kern-.08em
    T\kern-.1667em\lower.7ex\hbox{E}\kern-.125emX}}
\begin{document}

\title{Benchmarking the Effectiveness of Classification Algorithms and SVM Kernels for Dry Beans\\
{\footnotesize \textsuperscript{}}
\thanks{*Corresponding Author: psengupta\_be20@thapar.edu}
}

\author{
\IEEEauthorblockN{ Anant Mehta}
\IEEEauthorblockA{\textit{Computer Science and Engineering} \\
TIET, Patiala, India \\
amehta1\_be20@thapar.edu}
\and
\IEEEauthorblockN{*Prajit Sengupta}
\IEEEauthorblockA{\textit{Computer Science and Engineering} \\
TIET, Patiala, India \\
psengupta\_be20@thapar.edu}
\and
\IEEEauthorblockN{Divisha Garg}
\IEEEauthorblockA{\textit{Computer Science and Engineering} \\
TIET, Patiala, India \\
dgarg\_phd21@thapar.edu}
\and 
\IEEEauthorblockN{Harpreet Singh}
\IEEEauthorblockA{\textit{Faculty of Engineering} \\
Tel Aviv University, Israel \\
akalharpreet@gmail.com}
\and
\IEEEauthorblockN{Yosi Shacham Diamand}
\IEEEauthorblockA{\textit{Faculty of Engineering} \\
Tel Aviv University, Israel \\
yosish@tauex.tau.ac.il}
}
\maketitle

\begin{abstract}
Plant breeders and agricultural researchers can increase crop productivity by identifying desirable features, disease resistance, and nutritional content by analysing the Dry Bean dataset. This study analyses and compares different Support Vector Machine (SVM) classification algorithms, namely linear, polynomial, and radial basis function (RBF), along with other popular classification algorithms. The analysis is performed on the Dry Bean Dataset, with PCA (Principal Component Analysis) conducted as a preprocessing step for dimensionality reduction. The primary evaluation metric used is accuracy, and the RBF SVM kernel algorithm achieves the highest Accuracy of 93.34\%, Precision of 92.61\%, Recall of 92.35\% and F1 Score as 91.40\%. Along with adept visualization and empirical analysis, this study offers valuable guidance by emphasizing the importance of considering different SVM algorithms for complex and non-linear structured datasets.
\end{abstract}


\section{Introduction}
Dry Bean (DB) holds market importance in the agriculture, food processing, export-import, research, and consumer domains. It enables stakeholders to optimize production, ensure quality control, enhance market competitiveness, drive innovation, and meet consumer demands. By leveraging the available dataset of DB, market players can make informed decisions, establish more vital market positions, and contribute to the sustainable growth of the DB industry. Applying Machine Learning (ML) techniques to the DB dataset offers opportunities to enhance classification accuracy, improve quality assessment processes, predict crop yields, gain insights into bean characteristics, and automate various agricultural tasks. These advancements can positively impact the agricultural industry, contributing to increased productivity, improved quality control, and informed decision-making.
In the field of ML, classification algorithms play a vital role in discovering patterns and grouping similar data points together. An effective framework for classification and classification tasks is provided by SVM. Different SVM versions, as shown in Table \ref{tab1}, such as linear, polynomial, and RBF SVM, have been thoroughly explored and contrasted with other well-known classification algorithms in the context of classification.\cite{patle2013svm}.

The simplest and most used kernel in SVM is the linear kernel. It functions well for linearly separable data and expects a linear decision boundary. The polynomial kernel maps the data into a higher-dimensional space, allowing for non-linear decision boundaries. The RBF kernel is a well-liked option for processing extremely non-linear data, and thus, we explored all three kernel functions on the dataset\cite{barakat2010rule}. 

Prior to using these algorithms, PCA was carried out to lower the dataset's dimensionality. By transforming the data into a lower-dimensional space while preserving crucial information, PCA helps to mitigate the effects of dimensionality.\cite{daffertshofer2004pca}.

Support Vector Machines (SVM) is a powerful supervised learning algorithm used for classification and regression tasks. It finds an optimal hyperplane that separates data into different classes, maximizing the margin between them. SVM can handle both linearly separable and non-linearly separable data through the use of kernel functions.

The RBF SVM exhibited the highest accuracy of 93.34\% among the various SVM classification algorithms evaluated. This suggests that the this kernel effectively captured the underlying non-linear structures in the data, leading to improved classification performance\cite{fu2010mixing}. 
In the subsequent sections of this paper, we have discussed the experimental methodology, dataset acquisition, the selection of SVM kernels and empirical analysis of the classification algorithms.

\section{Related Work}
In 2023, Krishnan et al. employed machine learning methods to categorise the seven different types of dried beans using a multiclass classification model. Without balancing the dataset, Random Forest (RF) used by the authors achieved an accuracy of 92.06\%, while CatBoost achieved 92.76\% accuracy\cite{krishnan2023identification}.
Hasan et al. studied several varieties of DB before using a deep learning method to automatically categorise the beans. Their findings indicated that the method had an accuracy of 93.44\%\cite{hasan2021deep}. Dogan et al. in 2023 recommended SSA-ELM model, which had a 91.43\% success rate in classifying 14 different varieties of dry beans. Comparable findings showed that the suggested hybrid model outperformed conventional ML techniques in terms of classification accuracy and performance measures\cite{dogan2023dry}. In 2020, K-Nearest Neighbours (kNN), Decision Tree (DT), SVM and Multilayer Perceptron (MLP) classification models were developed using 10-fold cross validation strategy by Koklu et al. For MLP, SVM, KNN and DT, the overall accurate classification rates were found to be 91.73\%, 93.13\%, 87.92\%, and 92.52\%, respectively.\cite{koklu2020multiclass}.
Słowiński et al. visualised and analysed the DB dataset with various ML techniques that obtained results in the range of 88.35 – 93.61\%. The authors also compared different SVM classifiers with different hyperparameters. RBF along with certain set parameters achieved highest accuracy of 92.18\%\cite{slowinski2021dry}.
Using red-green-blue (RGB) pictures and ML methods, Pozzo et al. assessed computer vision in 2022 to identify and categorise different DB seed-borne fungus and their varieties. The authors employed a number of models, including SVM, Naive Bayes, RF, rpart, rpart1SE, and rpart2. The best score was obtained by RF with a score of around 80\%\cite{pozza2022using}.

 \begin{figure}[h]
    \centerline{\includegraphics[scale=0.8]{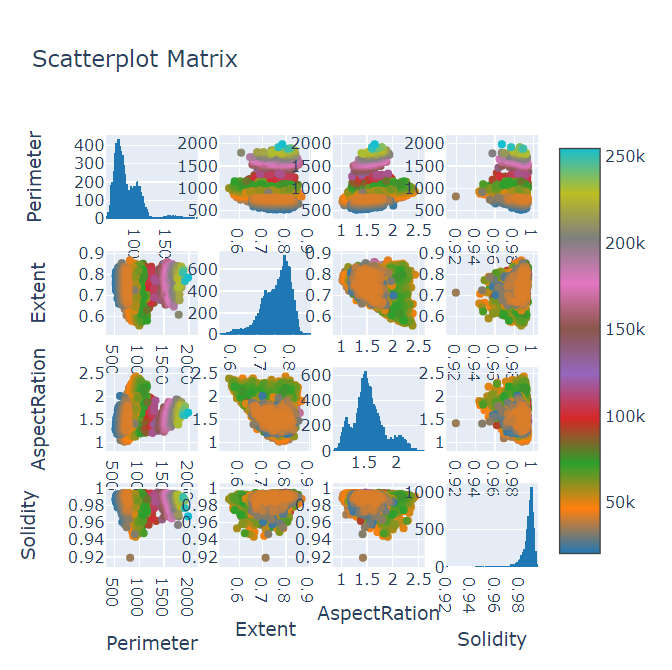}}
    \caption{Scatter Plot}
    \label{fig1}
    \end{figure}

\section{Methodology}
 \begin{figure*}[ht]
    \centerline{\includegraphics[scale=0.58]{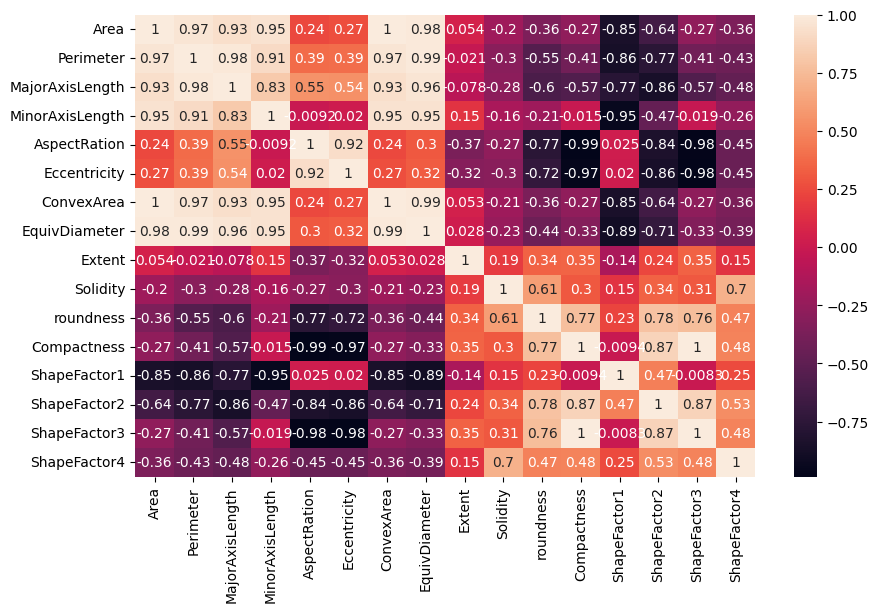}}
    \caption{Correlation Matrix}
    \label{fig2}
    \end{figure*}
\subsection{Dataset}\label{AA}
A common dataset in agricultural research and ML is the DB Dataset available on UCI ML Repository. For classification and pattern recognition applications, it offers useful details about numerous characteristics of several varieties of DB. \cite{koklu2020multiclass}. The collection includes 13 properties with associated class labels that explain various aspects of DB. The bean varieties represented by the class labels "BOMBAY", "SEKER", "DERMASON", "BARBUNYA", "CALI", "HOROZ" and "SIRA" are those that were included in the dataset.  The Scatter Plot for the different features has been shown in Figure \ref{fig1}. These labels enable supervised learning techniques, enabling the creation of models that can categorise fresh instances of dry beans according to their characteristics\cite{koklu2020multiclass}.
\begin{table}[h]
    \centering
    \caption{Types Of SVM}
    \label{tabsvm}
    \begin{tabular}{cccccccccc}
    \hline
    SVM Type & Parameter C & Parameter (Gamma) & Accuracy  \\
    \hline
    Linear & 1 & 0.0999 & 0.9280 \\
    Polynomial & 1 & 0.0999 & 0.9065 \\
    \textbf{RBF} & 1 & 0.0999 & \textbf{0.9334}  \\
    \hline
    \end{tabular}
    \label{tab1}
\end{table}

\subsection{Data Preprocessing}
\begin{itemize}
    \item \textbf{Data Visualization and Correlation Analysis:} Data visualization is a powerful tool that allows us to explore and understand complex datasets\cite{waskom2021seaborn} visually. By representing data graphically, patterns, trends, and correlations become more apparent, aiding in decision-making processes. Correlation analysis, on the other hand, helps to quantify and determine the strength and direction of relationships between variables\cite{hall1999correlation}. When combined, data visualization and correlation analysis provide invaluable insights into the interplay and dependencies within datasets, enabling us to confidently uncover meaningful insights and make data-driven decisions. Figure \ref{fig2} shows the correlation matrix of the dataset. It can be seen that features like Area, Perimeter, Convex Area, Major and Minor Axis length are quite related. Also, Figure \ref{fig5} shows the Anderson Plot of the different beans.

     \begin{figure}[h]
    \centerline{\includegraphics[scale=0.6]{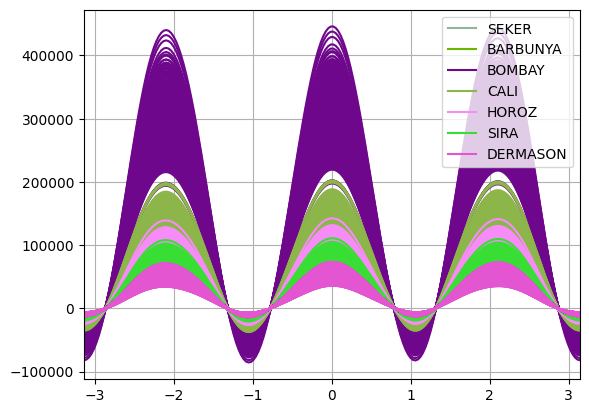}}
    \caption{Anderson Plot}
    \label{fig5}
    \end{figure}

    \item \textbf{Feature Reduction:}   
    The most crucial features can be extracted from a high-dimensional dataset using the dimensionality reduction approach known as PCA. PCA helps to preserve the most crucial information while discarding less relevant ones\cite{martinez2001pca}. 
    This reduction in dimensionality not only simplifies data visualization but also improves computational efficiency and mitigates the risk of overfitting in ML models\cite{mackiewicz1993principal}. The dataset here was reduced to 8 features (not correlated to each other), before feeding it for multiclass classification. Before applying PCA, the data was standardized. In Table \ref{tab2} classes 'SEKER', 'BARBUNYA', 'BOMBAY', 'CALI', 'HOROZ', 'SIRA' and 'DERMASON' map to label 5,0,1,2,4,6 and 3 respectively.
    
    \begin{table*}[htbp]
    \centering
    \caption{Pricipal Component Analysis}
    \label{tab:my-table}
    \begin{tabular}{cccccccccc}
    \hline
    S.No & F1 & F2 & F3 & F4 & F5 & F6 & F7 & F8 & Class \\
    \hline
    0 & -4.981561 & 1.824697 & 0.749021 & -0.390812 & -0.033531 & 0.301212 & 0.610269 & -0.274802 & 5.0 \\
    1 & -5.436792 & 2.932365 & 2.182374 & -0.431960 & 1.226464 & 0.045575 & 1.691342 & -0.317654 & 5.0 \\
    2 & -4.758088 & 1.826884 & 0.514038 & -0.125854 & 0.131506 & 0.208538 & 0.599563 & -0.162090 & 5.0 \\
    3 & -4.300541 & 2.003661 & 3.554447 & 0.082964 & 0.800766 & 0.502323 & 0.659708 & -0.672622 & 5.0 \\
    4 & -6.349340 & 4.088205 & 1.179199 & -0.830357 & -0.037073 & -0.278306 & 1.728546 & -0.893425 & 5.0 \\
    ... & ... & ... & ... & ... & ... & ... & ... & ... \\
    13606 & -1.125616 & -0.441079 & -0.875509 & -0.719279 & -0.298148 & 0.026482 & -0.484381 & -0.104817 & 3.0 \\
    13607 & -1.605011 & 0.495997 & -0.840558 & 0.797433 & 0.017084 & -0.090453 & -0.321913 & -0.034253 & 3.0 \\
    13608 & -1.417515 & 0.141194 & -0.387206 & -0.486439 & -0.383555 & -0.137333 & -0.333378 & 0.037884 & 3.0 \\
    13609 & -1.114666 & -0.212679 & 0.144088 & -0.841903 & -0.486805 & -0.097160 & -0.358602 & 0.034839 & 3.0 \\
    13610 & -0.766437 & -0.646514 & -0.994122 & 0.814679 & 0.258258 & 0.052163 & -0.295865 & -0.047087 & 3.0 \\
    \hline
    \end{tabular}
    \label{tab2}
    \end{table*}

    In PCA, a feature matrix is transformed using below equation.
     \begin{equation} \label{eq1}
     \gamma= \alpha*\beta
    \end{equation}
     where $\gamma$, $\alpha$, and $\beta$ represent the transformed data matrix with principal components, the original data matrix, and the matrix of eigenvectors (principal components), respectively. 

\end{itemize}

\section{Analysis of Proposed Work}
\subsection{Classification Algorithms}
In the context of the DB Dataset, various classification algorithms have been compared, including RBF, linear, and polynomial SVM. SVM is a potent method used for both regression and classification applications. It seeks to identify an ideal hyperplane that maximally distinguishes across classes.

\begin{itemize}
    \item \textbf{Linear SVM:} Linear SVM uses a linear kernel and assumes that the data can be separated by a straight line or hyperplane.
    It is suitable for datasets where the classes are well separated and can be linearly separated.
    Large-scale datasets may be used successfully with linear SVM because of their processing efficiency.\cite{soman2009machine}. Equation \ref{f1} shows the formula for 'linear' kernel:
    \begin{equation} \label{f1}
        \omega(x) = \text{k}(\mathbf{\zeta} \cdot \mathbf{x} + b)
    \end{equation}
    In the above formula, k is a constant that can be positive as well as negative. Term $b$ is constant, $\zeta$ refers to the weight vector and x refers to the input feature vector.
    \item \textbf{Polynomial SVM:} With polynomial SVM, the data is mapped into a higher-dimensional feature space using a polynomial kernel.
    It can capture nonlinear relationships by introducing polynomial terms.
    The degree of the polynomial determines the complexity of the decision boundary\cite{hussain2011comparison}. The formula for 'polynomial' kernel is shown below in equation \ref{f2}:
    \begin{equation}\label{f2}
        \omega(\mathbf{x}, \mathbf{y}) = (\zeta \cdot \mathbf{x} \cdot \mathbf{y} + c)^d
    \end{equation}
    Here, $d$ is the degree of polynomial and $c$ is the constant.
    \item \textbf{RBF SVM:} RBF SVM employs a Gaussian kernel, also referred to as an RBF kernel. It transforms the data into an infinite-dimensional feature space. RBF SVM can capture complex and nonlinear relationships in the data\cite{han2012parameter}.
     \begin{equation}\label{f3}
        \omega(\mathbf{x}, \mathbf{y}) = \exp\left(-\frac{{\|\mathbf{x} - \mathbf{y}\|^2}}{{2\sigma^2}}\right)
    \end{equation}
    Equation \ref{f3} (exponential curve) depicts the mathematical representation for the 'RBF' kernel, where $\sigma$ refers to the range/bandwidth of the kernel function.
    
\end{itemize}

When comparing these SVM variants with other classification algorithms\cite{hossin2015review} on the DB Dataset, it would be important to consider their performance in terms of accuracy, Matthews- correlation-coefficient (MCC), recall, precision, F1-score, kappa, area under the curve (AUC), and execution time as presented in Table \ref{tab:model-performance}. Each algorithm has its strengths and weaknesses, and the choice depends on the nature of the dataset and the specific requirements of the classification task.

\begin{table*}[htbp]
\centering
\caption{Model Performance}
\label{tab:model-performance}
\begin{tabular}{llcccccccc}
\hline
Model & Algorithm & Accuracy & AUC & Recall & Precision & F1 & Kappa & MCC & TT (Sec) \\
\hline
rf & Random Forest Classifier & 0.9261 & 0.9926 & 0.9261 & 0.9265 & 0.9261 & 0.9106 & 0.9107 & 2.3720 \\
xgboost & Extreme Gradient Boosting & 0.9253 & 0.9933 & 0.9253 & 0.9258 & 0.9253 & 0.9096 & 0.9097 & 8.8860 \\
ridge & Ridge Classifier & 0.8501 & 0.0000 & 0.8501&	0.8537&	0.8460 &	0.8175 &	0.8207 &	0.1100\\
ada	&Ada Boost Classifier&	0.4705	&0.7384	& 0.4705	&0.3392&0.3539 & 0.3275 &	0.3737&	0.9100\\
lightgbm & Light Gradient Boosting Machine & 0.9234 & 0.9934 & 0.9234 & 0.9238 & 0.9234 & 0.9073 & 0.9074 & 2.1010 \\
knn & K Neighbors Classifier & 0.9205 & 0.9828 & 0.9205 & 0.9213 & 0.9207 & 0.9039 & 0.9039 & 0.1270 \\
gbc & Gradient Boosting Classifier & 0.9196 & 0.9930 & 0.9196 & 0.9203 & 0.9196 & 0.9027 & 0.9029 & 20.9110 \\
et & Extra Trees Classifier & 0.9262 & 0.9932 & 0.9262 & 0.9270 & 0.9263 & 0.9107 & 0.9109 & 1.2460 \\
qda & Quadratic Discriminant Analysis & 0.9158 & 0.9916 & 0.9158 & 0.9180 & 0.9160 & 0.8984 & 0.8988 & 0.0800 \\
lda & Linear Discriminant Analysis & 0.8978 & 0.9909 & 0.8978 & 0.9107 & 0.8999 & 0.8765 & 0.8789 & 0.0830 \\
nb & Naive Bayes & 0.8964 & 0.9874 & 0.8964 & 0.9032 & 0.8974 & 0.8750 & 0.8762 & 0.0670 \\
lr & Logistic Regression & 0.9244 & 0.9934 & 0.9244 & 0.9252 & 0.9246 & 0.9086 & 0.9087 & 1.1810 \\
dt & Decision Tree Classifier & 0.8898 & 0.9322 & 0.8898 & 0.8902 & 0.8898 & 0.8668 & 0.8669 & 0.1470 \\
dummy	&Dummy Classifier &	0.2605 &	0.5000 &	0.2605 &	0.0679 &	0.1077 &	0.0000 &	0.0000 &	0.0690 \\
\hline
\end{tabular}
\end{table*}

\begin{figure}[ht]
    \centerline{\includegraphics[scale=0.45]{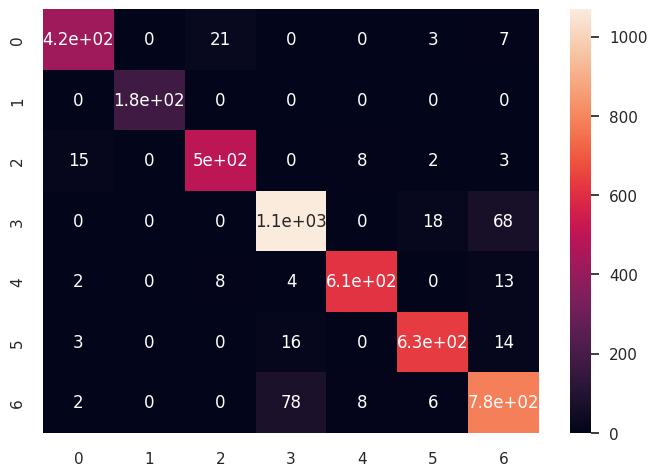}}
    \caption{Confusion Matrix}
    \label{fig3}
    \end{figure}
    
 \begin{figure}[ht]
    \centerline{\includegraphics[scale=0.45]{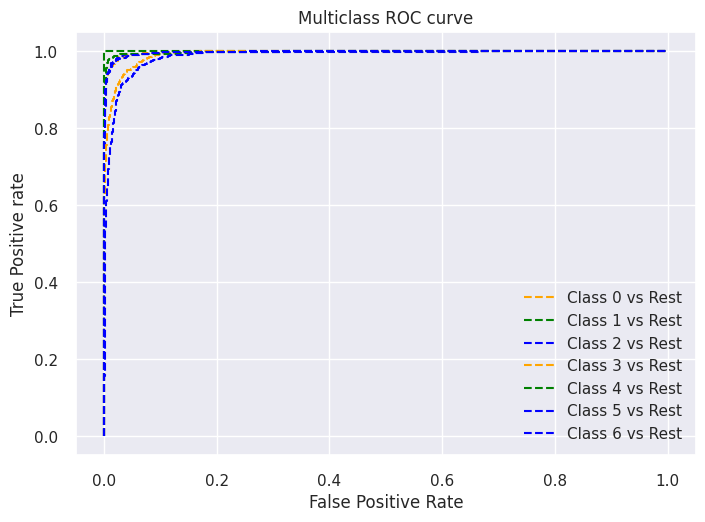}}
    \caption{ROC Curve}
    \label{fig4}
    \end{figure}

\section{Results}
When applying the RBF SVM algorithm to the DB dataset, the results revealed promising outcomes. This algorithm along with the set parameters outperformed other models explored in this study as well as other literature studies. The RBF SVM gave an accuracy of 93.34\% on the multiclass classification task. The confusion matrix for the task is shown in Figure \ref{fig3}. Other evaluation metrics used were Precision, Recall and F1 Score, giving scores of 92.61\%, 92.35\% and 91.40\%, respectively as shown in Table \ref{tab4}.
To further ensure the efficacy of the used model, Kappa\cite{chicco2020advantages} score was calculated as 0.895.\\
The performance of a classification model is depicted graphically by the ROC (Receiver Operating Characteristic) curve.\cite{hoo2017roc}. The ratio of true-positive rate to false-positive rate at various categorization criteria allows for evaluating and comparing different models based on their trade-off between sensitivity and specificity. The ROC curve for the seven classes is depicted in Figure \ref{fig4}.
The formulas for the used metrics are given below:
 \begin{equation} \label{eq2}
    Precision = \alpha / (\alpha + \beta)
    \end{equation}
  \begin{equation} \label{eq3}
    Recall = \alpha / (\alpha + \gamma)
    \end{equation} 
   where $\gamma$ is the count of 'false negatives', $\beta$ is the count of 'false positives', $\alpha$ is the number of 'true positives'\cite{hossin2015review}.\\
 \begin{equation} \label{eq4}
    F1 = 2 * (Precision * Recall) / (Precision + Recall)
    \end{equation} 
\begin{equation} \label{eq6}
        Kappa = (OA - EA)\\ / (1 - EA)  
    \end{equation}
  where observed accuracy (OA) is the count of legitimate predictions, and expected accuracy (EA) is the number of legitimate predictions expected by chance. \\

\begin{table}[]
\centering
\caption{Comparison of Results}
\label{tab4}
\begin{tabular}{|l|l|}
\hline
\textbf{Model}                & \textbf{Accuracy} \\ \hline
Random Forest\cite{krishnan2023identification}                & 92.06\%           \\ \hline
CatBoost \cite{krishnan2023identification}                     & 92.76\%           \\ \hline
Deep Neural Network \cite{hasan2021deep}          & 93.44\%           \\ \hline
SSA-ELM   \cite{dogan2023dry}                    & 91.43\%           \\ \hline
rpart1 SE, rpart2 Naive Bayes \cite{pozza2022using}  & 80.00\%        \\ \hline
\textbf{SVM RBF Kernel}       & \textbf{93.34\%}  \\ \hline
\end{tabular}
\end{table}

\section{Experimental Setup}
A CPU with 16 GB of RAM was used to run these models. Basic data science tools including NumPy, Pandas, Scikit-Learn, and PyCaret were used to write the code. Python 3.0 was the main language employed.
\section{Future Scope \& Discussion}
In several artificial intelligence applications, SVM with the RBF kernel have demonstrated promising results. When applied to the DB dataset, SVM RBF exhibits a vast future scope for analysis and discussion. Several areas of potential exploration and discussion concerning SVMs and other ML algorithms on the DB dataset exist. Researchers can investigate the impact of different kernel parameters on the model's performance by exploring the optimal kernel width and regularization parameter values. Additionally, feature selection techniques can be employed to determine the most informative attributes for classification. Exploring ensemble methods, such as combining multiple
ML models using boosting and bagging approaches. The interpretability of deep learning models is another topic for discussion, as understanding the decision-making process can help gain insights into the features and patterns that contribute to the classification. Furthermore, exploring methods to mitigate overfitting, handle class imbalance, and improve generalization on limited data can be crucial research areas.

\bibliographystyle{unsrt}
\bibliography{bibfile}

\begin{thebibliography}{10}

\bibitem{patle2013svm}
Arti Patle and Deepak~Singh Chouhan.
\newblock Svm kernel functions for classification.
\newblock In {\em 2013 International Conference on Advances in Technology and
  Engineering (ICATE)}, pages 1--9. IEEE, 2013.

\bibitem{barakat2010rule}
Nahla Barakat and Andrew~P Bradley.
\newblock Rule extraction from support vector machines: a review.
\newblock {\em Neurocomputing}, 74(1-3):178--190, 2010.

\bibitem{daffertshofer2004pca}
Andreas Daffertshofer, Claudine~JC Lamoth, Onno~G Meijer, and Peter~J Beek.
\newblock Pca in studying coordination and variability: a tutorial.
\newblock {\em Clinical biomechanics}, 19(4):415--428, 2004.

\bibitem{fu2010mixing}
Zhouyu Fu, Antonio Robles-Kelly, and Jun Zhou.
\newblock Mixing linear svms for nonlinear classification.
\newblock {\em IEEE Transactions on Neural Networks}, 21(12):1963--1975, 2010.

\bibitem{krishnan2023identification}
S~Krishnan, SK~Aruna, Karthick Kanagarathinam, and Ellappan Venugopal.
\newblock Identification of dry bean varieties based on multiple attributes
  using catboost machine learning algorithm.
\newblock {\em Scientific Programming}, 2023:1--21, 2023.

\bibitem{hasan2021deep}
Md~Mahadi Hasan, Muhammad~Usama Islam, and Muhammad~Jafar Sadeq.
\newblock A deep neural network for multi-class dry beans classification.
\newblock In {\em 2021 24th International Conference on Computer and
  Information Technology (ICCIT)}, pages 1--5. IEEE, 2021.

\bibitem{dogan2023dry}
Musa Dogan, Yavuz~Selim Taspinar, Ilkay Cinar, Ramazan Kursun, Ilker~Ali Ozkan,
  and Murat Koklu.
\newblock Dry bean cultivars classification using deep cnn features and salp
  swarm algorithm based extreme learning machine.
\newblock {\em Computers and Electronics in Agriculture}, 204:107575, 2023.

\bibitem{koklu2020multiclass}
Murat Koklu and Ilker~Ali Ozkan.
\newblock Multiclass classification of dry beans using computer vision and
  machine learning techniques.
\newblock {\em Computers and Electronics in Agriculture}, 174:105507, 2020.

\bibitem{slowinski2021dry}
Grzegorz S{\l}owi{\'n}ski.
\newblock Dry beans classification using machine learning.
\newblock {\em Proceedings http://ceur-ws. org ISSN}, 1613:0073, 2021.

\bibitem{pozza2022using}
Edson~Amp{\'e}lio Pozza, Marcelo de~Carvalho~Alves, and Luciana Sanches.
\newblock Using computer vision to identify seed-borne fungi and other targets
  associated with common bean seeds based on red--green--blue spectral data.
\newblock {\em Tropical Plant Pathology}, pages 1--18, 2022.

\bibitem{waskom2021seaborn}
Michael~L Waskom.
\newblock Seaborn: statistical data visualization.
\newblock {\em Journal of Open Source Software}, 6(60):3021, 2021.

\bibitem{hall1999correlation}
Mark~A Hall.
\newblock {\em Correlation-based feature selection for machine learning}.
\newblock PhD thesis, The University of Waikato, 1999.

\bibitem{martinez2001pca}
Aleix~M Martinez and Avinash~C Kak.
\newblock Pca versus lda.
\newblock {\em IEEE transactions on pattern analysis and machine intelligence},
  23(2):228--233, 2001.

\bibitem{mackiewicz1993principal}
Andrzej Ma{\'c}kiewicz and Waldemar Ratajczak.
\newblock Principal components analysis (pca).
\newblock {\em Computers \& Geosciences}, 19(3):303--342, 1993.

\bibitem{soman2009machine}
KP~Soman, R~Loganathan, and V~Ajay.
\newblock {\em Machine learning with SVM and other kernel methods}.
\newblock PHI Learning Pvt. Ltd., 2009.

\bibitem{hussain2011comparison}
Muhammad Hussain, Summrina~Kanwal Wajid, Ali Elzaart, and Mohammed Berbar.
\newblock A comparison of svm kernel functions for breast cancer detection.
\newblock In {\em 2011 eighth international conference computer graphics,
  imaging and visualization}, pages 145--150. IEEE, 2011.

\bibitem{han2012parameter}
Shunjie Han, Cao Qubo, and Han Meng.
\newblock Parameter selection in svm with rbf kernel function.
\newblock In {\em World Automation Congress 2012}, pages 1--4. IEEE, 2012.

\bibitem{hossin2015review}
Mohammad Hossin and Md~Nasir Sulaiman.
\newblock A review on evaluation metrics for data classification evaluations.
\newblock {\em International journal of data mining \& knowledge management
  process}, 5(2):1, 2015.

\bibitem{chicco2020advantages}
Davide Chicco and Giuseppe Jurman.
\newblock The advantages of the matthews correlation coefficient (mcc) over f1
  score and accuracy in binary classification evaluation.
\newblock {\em BMC genomics}, 21:1--13, 2020.

\bibitem{hoo2017roc}
Zhe~Hui Hoo, Jane Candlish, and Dawn Teare.
\newblock What is an roc curve?, 2017.

\end{thebibliography}
\vspace{12pt}

\end{document}